\documentclass[10pt,twocolumn,letterpaper]{article}

\usepackage{cvpr}              

\usepackage{xcolor}
\definecolor{cvprblue}{rgb}{0.21,0.49,0.74}
\usepackage[pagebackref,breaklinks,colorlinks,allcolors=cvprblue]{hyperref}
\usepackage{algorithm}
\usepackage{algpseudocode}

\def\paperID{25} 
\def\confName{CVPR}
\def\confYear{2026}

\title{Deployment-Aligned Low-Precision Neural Architecture Search for Spaceborne Edge AI}

\author{
Parampuneet Kaur Thind$^{1,2,4}$,
Vaibhav Katturu$^{3}$,
Giacomo Zema$^{4}$,
Roberto Del Prete$^{2}$\\
$^{1}$Sapienza University of Rome\\
$^{2}$ $\Phi$-lab, ESA ESRIN\\
$^{3}$University of Pisa\\
$^{4}$Argotec Srl\\
{\tt\small parampuneet.thind@uniroma1.it}\\
{\tt\small v.katturu@studenti.unipi.it}\\
{\tt\small giacomo.zema@argotecgroup.com}\\
{\tt\small roberto.delprete@esa.int}
}
\begin{document}
\maketitle
\begin{abstract}

Designing deep networks that meet strict latency and accuracy constraints on edge accelerators increasingly relies on hardware-aware optimization, including neural architecture search (NAS) guided by device-level metrics. Yet most hardware-aware NAS pipelines still optimize architectures under full-precision assumptions and apply low-precision adaptation only after the search, leading to a mismatch between optimization-time behavior and deployment-time execution on low-precision hardware that can substantially degrade accuracy.
We address this limitation by integrating deployment-aligned low-precision training directly into hardware-aware NAS. Candidate architectures are exposed to FP16 numerical constraints during fine-tuning and evaluation, enabling joint optimization of architectural efficiency and numerical robustness without modifying the search space or evolutionary strategy.
We evaluate the proposed framework on vessel segmentation for spaceborne maritime monitoring, targeting the Intel\textregistered{} Movidius\texttrademark{} Myriad\texttrademark{} X Visual Processing Unit (VPU). While post-training precision conversion reduces on-device performance from 0.85 to 0.78 mIoU, deployment-aligned low-precision training achieves 0.826 mIoU on-device for the same architecture (95{,}791 parameters), recovering approximately two-thirds of deployment-induced accuracy gap without increasing model complexity.
These results demonstrate that incorporating deployment-consistent numerical constraints into hardware-aware NAS substantially improves robustness and alignment between optimization and deployment for resource-constrained edge Artificial Intelligence (AI).

\end{abstract}

\section{Introduction}
\label{sec:intro}

Neural Architecture Search (NAS) has emerged as a powerful paradigm for automatically designing deep neural networks that balance accuracy and efficiency under resource constraints \cite{POYSER2024110052}. Hardware-aware NAS extends this idea by incorporating device-level performance metrics such as latency, throughput, or memory footprint directly into the optimization loop, enabling architectures to be selected with deployment constraints in mind \cite{DelPrete2025NAS, Thind2025}. However, despite explicitly accounting for hardware characteristics, hardware-aware NAS pipelines still optimize candidate architectures under full-precision floating-point (FP32) training assumptions and apply low-precision adaptation only after the search is complete \cite{dong2018dppnetdeviceawareprogressivesearch}. This decoupling introduces a systematic mismatch between optimization-time behavior and deployment-time execution on low-precision edge accelerators, often resulting in substantial accuracy degradation once models are deployed.

In this work, we address this limitation by integrating deployment-aligned low-precision effects directly into the hardware-aware NAS evaluation loop. Rather than treating reduced numerical precision as a post-training concern, candidate architectures are exposed to low-precision floating point (FP16) numerical constraints during fine-tuning and on-device evaluation. This allows the NAS process to jointly optimize architectural efficiency and numerical robustness without modifying the search space or the underlying evolutionary strategy.

This problem is particularly relevant in Earth Observation (EO), where missions increasingly explore on-board data processing to enable faster and more autonomous decision-making in spaceborne systems \cite{thind2026, Pastena2020, Melega2025}. Applications such as maritime monitoring \cite{10283401}, disaster response \cite{TRABA2026105095, mateo2021towards, del2021board}, and environmental surveillance \cite{UNSPIDER2025} benefit from reducing the latency between image acquisition and event detection. Traditional EO pipelines rely on downlinking raw or minimally processed imagery to ground stations, introducing delays due to limited visibility windows, bandwidth constraints, and the need for multiple ground station passes before actionable information can be generated \cite{zhang2022progress}.

Recent missions and demonstrators have shown that performing inference directly on-board can significantly reduce these delays by transmitting extracted information instead of raw data \cite{Melega2023}. However, this shift toward on-board intelligence places stringent constraints on the computational resources available for data processing \cite{Furano2020}. Spaceborne platforms particularly small satellites and CubeSats operate under tight limits on power consumption, memory footprint, and processing throughput, requiring deep learning models that are both compact and efficient \cite{Cratere2024}.

Designing neural networks that satisfy these constraints while maintaining task performance remains challenging. Simply reducing model size or complexity does not reliably preserve accuracy, especially for dense prediction tasks \cite{Ajmoud}. Moreover, space missions increasingly rely on heterogeneous commercial off-the-shelf edge accelerators, which exhibit distinct architectural and numerical characteristics despite operating within similar power envelopes \cite{Furano2020}. As a result, models trained and optimized under conventional full-precision settings are particularly vulnerable to deployment-time performance degradation when executed under reduced numerical precision.

Hardware-aware NAS has been successfully applied to spaceborne inference, demonstrating favorable accuracy–efficiency trade-offs on edge platforms through device-in-the-loop evaluation \cite{DelPrete2025NAS}. Nevertheless, existing approaches rely on post-training precision conversion to adapt selected architectures to low-precision hardware, leaving numerical robustness largely unaddressed during the search itself \cite{cheng2018model}. This limitation motivates the need for NAS pipelines that explicitly align optimization-time assumptions with deployment-time numerical behavior.

We evaluate the proposed deployment-aligned training strategy on a vessel segmentation task for maritime monitoring, using the same dataset and experimental protocol as recent NAS-based EO studies \cite{Thind2025}. All candidate architectures are evaluated directly on an Intel\textregistered{} Movidius\texttrademark{} Myriad\texttrademark{} X Vision Processing Unit (VPU) accelerator, enabling a realistic assessment of segmentation accuracy, inference latency, and throughput under operational edge deployment conditions. The results show that incorporating deployment-consistent numerical constraints within the NAS loop substantially reduces the performance gap between Graphics Processing Unit (GPU)-based optimization and on-device inference, while maintaining compact model size and real-time execution capability. The main contributions of this work are summarized as follows:
\begin{itemize}
    \item \textbf{Deployment-aligned NAS:} We integrate low-precision-aware training directly into a hardware-aware NAS loop, aligning optimization-time behavior with deployment-time FP16 execution without modifying the search space or evolutionary strategy.
    \item \textbf{Search-level numerical alignment:} We show that incorporating deployment-consistent precision during evaluation changes architecture selection dynamics and reduces the optimization, deployment accuracy gap compared to post-training precision conversion.
    \item \textbf{Realistic device-in-the-loop validation:} We provide comprehensive on-device evaluation on an Intel\textregistered{} Movidius\texttrademark{} Myriad\texttrademark{} X VPU, measuring segmentation accuracy and throughput under real deployment conditions.
\end{itemize}

The remainder of this paper is organized as follows. Section~2 reviews related work on model compression, neural architecture search, and low-precision deployment. Section~3 presents the proposed hardware-aware NAS framework with deployment-aligned low-precision training. Section~4 reports quantitative and qualitative experimental results. Section~5 concludes the paper and discusses future research directions.



\section{Background and Related Work}

\subsection{On-Board Earth Observation and Edge Constraints}

Deploying deep learning models on-board spacecraft requires aggressive reductions in model size and computational footprint due to strict constraints on memory capacity, non-volatile storage, power consumption, and thermal dissipation imposed by on-board computers and edge accelerators \cite{Choudhary2020}. Space-qualified hardware platforms typically provide limited on-chip memory and restricted external storage, making it infeasible to deploy large high-capacity neural networks without careful optimization.

These constraints are particularly relevant for EO missions, where on-board processing is increasingly adopted to reduce downlink latency and allow timely decision-making \cite{tuia2024artificial}. However, the combination of dense prediction tasks, such as semantic segmentation, and heterogeneous commercial off-the-shelf edge accelerators poses significant challenges for reliable deployment. As a result, model efficiency and robustness have become central concerns in spaceborne AI pipelines.

\subsection{Hardware-Aware Neural Architecture Search}

Rather than compressing an already trained model, NAS aims to discover efficient neural network architectures from scratch by formulating architecture design as a constrained optimization problem on a predefined search space $\mathcal{A}$. The objective is to identify an architecture that minimizes task-specific loss while satisfying deployment-related constraints such as latency, memory footprint, or computational cost:
\begin{equation}
\min_{a \in \mathcal{A}} \; \mathcal{L}(a, \mathcal{D}) \quad \text{s.t.} \quad C(a) \leq C_{\text{max}}.
\end{equation}

Hardware-aware NAS extends this paradigm by incorporating device-level performance metrics directly into the optimization loop, enabling architectures to be selected with deployment constraints in mind \cite{dong2018dppnetdeviceawareprogressivesearch}. Early NAS approaches relied on population-based or reinforcement learning strategies \cite{Colin, Kaelbling1996}, while more recent methods improve efficiency through weight sharing, differentiable relaxation, or direct device-aware optimization \cite{zela2020, liu2019, cai2019}.

NAS has been successfully applied to remote sensing and EO tasks, including object recognition, classification, and onboard inference \cite{rs15010091, Liu2022PolNASAN, cassimon2024}. In particular, prior work has demonstrated the feasibility of embedding hardware evaluation within the NAS loop for spaceborne deployment, achieving favorable accuracy–efficiency trade-offs on edge accelerators \cite{DelPrete2025NAS}. However, most hardware-aware NAS pipelines optimize candidate architectures under full-precision training assumptions and adapt them to low-precision hardware only after the search is complete.

\subsection{Low-Precision Deployment and Quantization-Aware Training}

To enable efficient execution on resource-constrained hardware, model compression techniques such as quantization, pruning, knowledge distillation, and low-rank decomposition are commonly employed \cite{Choudhary2020}. Among these, quantization is particularly effective in reducing memory footprint, computational cost, and energy consumption by lowering the numerical precision of weights and activations \cite{nagel2021whitepaperneuralnetwork}.

Post-training quantization (PTQ) applies precision reduction after full-precision training, typically relying on calibration data to estimate quantization parameters \cite{banner2019, cai2020zeroqnovelzeroshot}. While PTQ enables rapid deployment, it is sensitive to numerical mismatch and often leads to accuracy degradation on compact architectures and low-bit precision hardware \cite{choukroun2019lowbitquantizationneuralnetworks}. 

Quantization-aware training (QAT) mitigates this issue by explicitly modeling low-precision effects during training using fake-quantization operators and straight-through estimators \cite{nagel2021whitepaperneuralnetwork}. QAT generally yields higher robustness at deployment time, but it is typically applied after architecture selection and treated as a separate post-optimization step \cite{cambier2020shiftedsqueezed8bitfloating, chmiel2020neuralgradientsnearlognormalimproved}.

\subsection{Limitations of Prior Work}

Despite advances in both hardware-aware NAS and quantization-aware training, these two lines of research have largely evolved independently. Existing approaches either (i) incorporate hardware metrics into NAS while optimizing architectures under full-precision assumptions, or (ii) apply low-precision training after the architecture has been selected. This separation introduces a mismatch between optimization-time behavior and deployment-time execution, particularly for spaceborne edge accelerators operating under reduced numerical precision.

This limitation motivates the integration of deployment-aware numerical constraints directly into the NAS process. By exposing candidate architectures to low-precision effects during evaluation, the search itself can favor architectures that are not only structurally efficient, but also numerically robust under real deployment conditions.


\section{Methodology}
\label{sec:methodology}

This work extends hardware-aware NAS by explicitly aligning the numerical conditions observed during optimization with those encountered at deployment time on edge accelerators. While existing hardware-aware NAS frameworks incorporate device-level performance metrics (e.g., latency or throughput) through post-training on-device evaluation, they typically rely on full-precision (FP32) training on GPUs and implicitly assume numerical equivalence at deployment. In practice, edge accelerators frequently execute reduced-precision arithmetic, introducing discrepancies in dynamic range, rounding behavior, and accumulation error that are not captured during FP32 optimization. 

Our key idea is to integrate deployment-aligned low-precision effects directly into the NAS evaluation loop. Instead of treating precision conversion as a post-selection adaptation step, candidate architectures are exposed to same numerical regime used at deployment during fine-tuning and on-device evaluation. This biases the search toward architectures that are not only structurally efficient, but also numerically robust under the target execution precision.

\begin{figure*}[t]
    \centering
    \includegraphics[width=\textwidth]{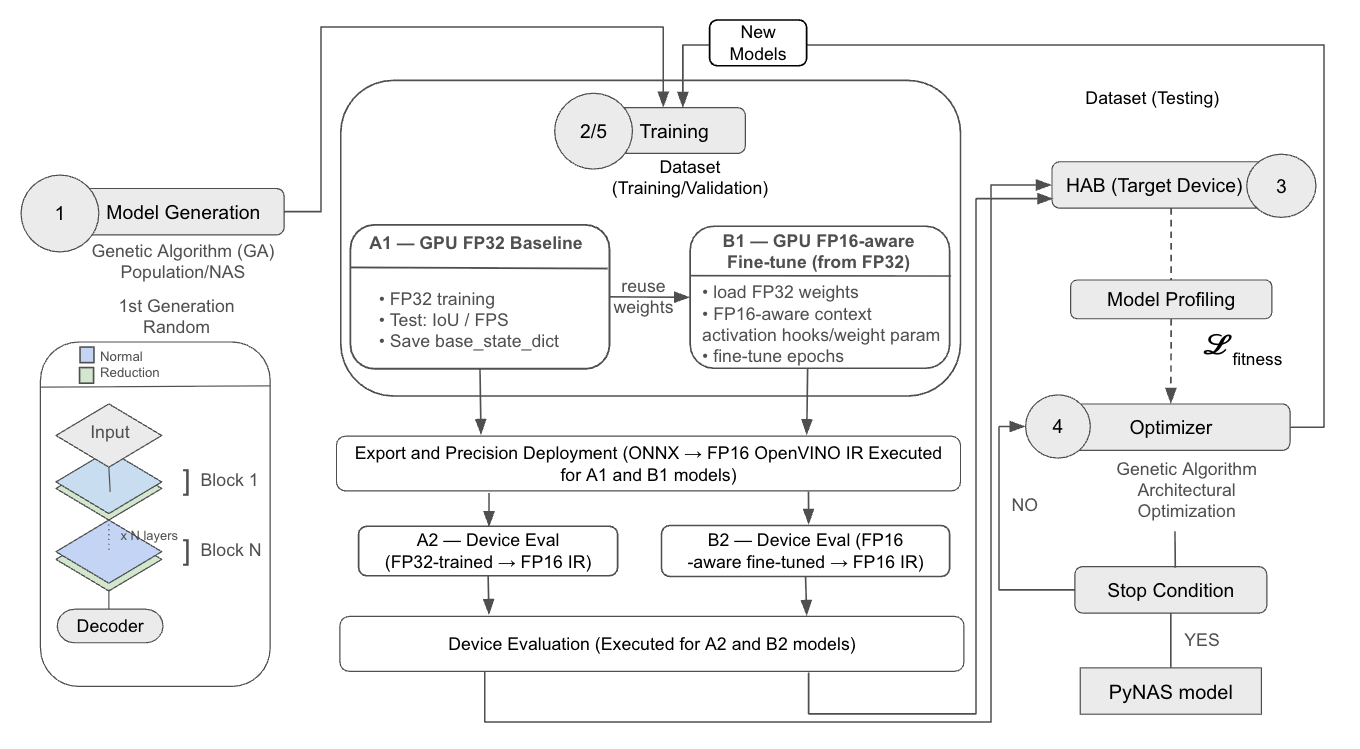}
    \caption{
    The diagram illustrates the proposed NAS framework in which each candidate architecture generated by the genetic algorithm
    is trained and evaluated under two deployment-aligned numerical regimes. First, an FP32 baseline model is trained on GPU (A1) and
    directly exported to FP16 OpenVINO intermediate representation, enabling post-training precision conversion (PTQ-style deployment)
    and subsequent on-device evaluation (A2). In parallel, the same architecture is fine-tuned using low-precision-aware training (B1), initial-
    ized from the FP32 weights and incorporating fake-FP16 activation rounding and weight parametrization with straight-through estimators,
    before being exported to FP16 OpenVINO IR and evaluated on the target hardware (B2). Both FP32-trained and FP16-aware fine-tuned
    models are therefore deployed and benchmarked on the target device, allowing a direct comparison between PTQ-based deployment and
    deployment-aligned low-precision training. The NAS fitness $\mathcal{L}$ is computed from on-device performance metrics, while the paired evalu-
    ation paths enable quantitative and qualitative assessment of accuracy degradation due to post-training conversion and the corresponding
    recovery achieved through low-precision-aware fine-tuning, as reported in the results.
    }
    \label{fig:nas-framework}
\end{figure*}

Figure~\ref{fig:nas-framework} summarizes the proposed workflow. For each candidate architecture sampled by the genetic algorithm, we evaluate two deployment scenarios: (i) \textbf{post-training precision conversion (PTQ baseline)}, where an FP32-trained model is exported to an FP16 OpenVINO intermediate representation (IR) and executed on the target device, and (ii) \textbf{deployment-aligned low-precision training}, where the same architecture is fine-tuned under FP16-aware projections prior to export and on-device evaluation. Importantly, both branches share the same architecture, search space, and evolutionary operators; only the evaluation pathway differs.

\subsection{Hardware-Aware Neural Architecture Search}

\paragraph{Search space:}
The admissible set of candidate architectures is defined as a discrete search space $\mathcal{A}$ composed of single-path networks with up to $M_n{=}6$ learnable blocks. Each block is sampled from a library of convolutional primitives and macro-blocks:
(i) convolutional layers with activation (\texttt{ConvAct}) or with batch normalization (\texttt{ConvBnAct}),
(ii) squeeze-and-excitation convolutional blocks (\texttt{ConvSE}),
(iii) mobile inverted bottleneck blocks with and without residual connections (\texttt{MBConv}, \texttt{MBConvNoRes}),
(iv) cross-stage partial variants (\texttt{CSPConvBlock}, \texttt{CSPMBConvBlock}),
(v) DenseNet-style blocks (\texttt{DenseNetBlock}),
(vi) ResNet-style blocks (\texttt{ResNetBlock}),
with optional pooling (\texttt{AvgPool}, \texttt{MaxPool}), dropout, and a fixed segmentation head.
Within each primitive, discrete and continuous hyperparameters are sampled from predefined ranges, including kernel size (1--5 for standard conv blocks), channel-width coefficients (e.g., 4--24 depending on block type), and activation functions (ReLU or GELU). An architecture is represented by a structured \emph{architecture code} (a compact genotype string) that deterministically parses into a valid network, enabling mutation and crossover while preserving structural validity.

\paragraph{Genetic algorithm settings:}
We use a population-based evolutionary NAS strategy with population size $P_s{=}16$ and $G{=}10$ generations. At each generation, candidates are ranked using a hardware-aware fitness function and updated through (i) elitism, retaining the top $k_{\text{best}}{=}1$ architectures, (ii) diversity injection, sampling $n_{\text{random}}{=}6$ new architectures uniformly from $\mathcal{A}$, and (iii) genetic operators applied to the mating pool consisting of the top 50\% of the population. Mutation is applied with probability $p_{\text{mut}}{=}0.15$, and single-point crossover recombines parent genotypes to generate offspring.

\paragraph{Fitness:}
The fitness of an architecture $a$ is computed as
\begin{equation}
\mathcal{F}(a) = \alpha \cdot \mathrm{fps}(a) + \beta \cdot \mathrm{Metric}(a)\cdot e^{\gamma \cdot \mathrm{Metric}(a)},
\end{equation}
where $\mathrm{fps}(a)$ is inference throughput measured on the target device and $\mathrm{Metric}(a)$ is task accuracy (mIoU). The exponential term increases sensitivity to high-accuracy solutions, biasing the search toward architectures that satisfy both predictive and deployment constraints.
 
\paragraph{Device-in-the-loop evaluation:}
After training (and optional fine-tuning, described below), each candidate architecture is exported to an FP16 OpenVINO intermediate representation (IR) and executed on the Intel\textregistered{} Movidius\texttrademark{} Myriad\texttrademark{} X VPU. Throughput and latency are measured directly on-device. This captures hardware-specific effects such as operator scheduling, memory access patterns, and reduced-precision arithmetic behavior that are not observable during GPU training. These measurements feed back into the NAS loop via $\mathcal{F}(a)$.

\subsection{Deployment-Aligned Low-Precision Training}

\paragraph{Training budget per candidate:}
Each candidate architecture is trained for $E_{\text{FP32}}{=}10$ epochs in FP32 using batch size 32. To model deployment-time reduced precision, we then perform FP16-aware fine-tuning for $E_{\text{LP}}{=}10$ epochs (after a 1-epoch warmup), initialized from the FP32-trained weights. This staged procedure isolates the effect of deployment-aligned numerical adaptation: the PTQ baseline and the FP16-aware model share the same initialization and architecture.

\paragraph{FP16 projection operator and affected layers:}
Rather than integer-only quantization, we model reduced-precision effects through FP16 projection consistent with the target deployment pipeline. Let $f_{\theta}$ denote a network with parameters $\theta$. During fine-tuning, we inject a projection operator $\mathcal{P}(\cdot)$ into the forward pass:
\begin{equation}
y = \mathcal{P}\big(f_{\theta}(x)\big),
\end{equation}
where $\mathcal{P}$ casts activations to FP16 and back to FP32 to simulate reduced mantissa precision and rounding. Projections are applied to the outputs of convolutional and linear operators (i.e., the primary accumulation sites), while gradients are accumulated in FP32. Weight rounding is enabled during the forward pass (\texttt{fp16\_round\_weights=true}) using a straight-through estimator (STE),
\begin{equation}
\frac{\partial \mathcal{P}(x)}{\partial x} \approx 1,
\end{equation}
so training remains stable while the forward computation reflects reduced precision.

\paragraph{Activation clipping:}
To better match the dynamic range observed under FP16 deployment and reduce overflow sensitivity, activations are clipped during FP16-aware fine-tuning using a fixed symmetric bound $|x| \le 12$. This threshold was selected empirically based on the typical activation distributions observed during training and reflects a trade-off between preserving dynamic range and avoiding FP16 overflow and rounding amplification effects \cite{PACT2018}.

\paragraph{FP16 projection for Myriad X deployment:}
Our deployment pipeline exports networks to an FP16 OpenVINO intermediate representation (IR) and executes them on the Intel\textregistered{} Movidius\texttrademark{} Myriad\texttrademark{} X VPU. FP16-aware projections therefore emulate the dominant numerical effect introduced by the deployment toolchain reduced mantissa precision and rounding in FP16 arithmetic while maintaining compatibility with standard FP32 optimizers. Although the Intel\textregistered{} Movidius\texttrademark{} Myriad\texttrademark{} X VPU may still exhibit additional device-specific behavior (e.g., operator fusion or internal accumulation details), using FP16 projections during fine-tuning directly targets the same precision regime used at inference and substantially reduces the optimization--deployment mismatch observed with PTQ.

\paragraph{Integration into NAS:}
For each candidate architecture, we first compute an on-device PTQ baseline by exporting the FP32-trained model to FP16 IR and evaluating it on the target device. We then fine-tune the same architecture under FP16-aware projections, re-export, and re-evaluate on-device. Fitness is computed from the deployment-aligned on-device metrics, guiding evolution toward architectures that are both hardware-efficient and numerically robust.


\section{Results}
\label{sec:results}

\subsection{Dataset}

All experiments are conducted using the HRSC2016 dataset, a publicly available benchmark for high-resolution ship detection and segmentation in optical satellite imagery \cite{HRSC2016}. The dataset consists of images primarily collected from Google Earth, with ground sampling distances ranging from approximately 0.4\,m to 2\,m. Image dimensions vary significantly, spanning from roughly $300 \times 300$ pixels to over $1500 \times 900$ pixels, with most samples exceeding $1000 \times 600$ pixels in size.

Original annotations are provided as oriented bounding boxes with associated vessel categories, covering both civilian and military ships across 25 classes. In this work, the task is formulated as binary vessel segmentation, where all ship categories are merged into a single foreground class and the remaining pixels are treated as background. This formulation reflects realistic onboard processing requirements, where timely and reliable vessel localization is often prioritized over fine-grained classification.

To obtain pixel-level supervision, bounding box annotations were converted into dense segmentation masks. Images were intensity-normalized and converted to a common format, after which annotation files were parsed to extract bounding box information. These bounding boxes were used to generate initial segmentation masks, which were subsequently refined through additional processing and manual inspection to ensure accurate vessel coverage. All samples were resized to $512 \times 512$ pixels using zero-padding when necessary and stored as NumPy arrays together with their corresponding masks for efficient training and evaluation.

Additional preprocessing steps were applied to improve dataset quality and robustness. In particular, targeted augmentation was used to address imbalances between small and large vessels, while corrupted or low-quality samples were removed. Extreme or ambiguous annotations were manually reviewed. This preprocessing pipeline results in a clean and diverse dataset suitable for evaluating both segmentation accuracy and deployment robustness under reduced numerical precision.

\subsection{Evaluation Metrics}

Segmentation performance is evaluated using mean Intersection-over-Union (mIoU), computed over all valid classes and averaged across the batch. The mIoU computation follows the same rules used during on-device evaluation, excluding classes with empty unions and ignoring invalid pixels when applicable ensuring consistency between GPU-based evaluation and deployment-time measurements. Architecture selection during NAS is driven by hardware-aware fitness function defined in Eq.~(3), which jointly accounts for segmentation accuracy and inference throughput measured directly on the target device.

\subsection{Experimental Setup}

All candidate architectures are trained on a workstation equipped with an AMD Radeon\texttrademark{} GPU using full-precision (FP32) arithmetic \cite{AMDRadeon}. Deployment-time evaluation is performed on the Intel\textregistered{} Movidius\texttrademark{} Myriad\texttrademark{} X, a low-power edge accelerator representative of onboard computing platforms used in spaceborne systems \cite{IntelMyriadX}.

Two experimental configurations are evaluated within the same hardware-aware NAS framework:

\begin{itemize}
    \item \textbf{Post-training precision conversion (baseline):} Architectures are trained exclusively in FP32 on the GPU. After training, models are exported and evaluated on the Intel\textregistered{} Movidius\texttrademark{} Myriad\texttrademark{} X VPU using post-training precision conversion, without exposure to reduced-precision effects during optimization.
    \item \textbf{Low-precision-aware fine-tuning:} Architectures are first trained in FP32 and subsequently fine-tuned with low-precision-aware training, where FP16 numerical effects are explicitly modeled during the forward pass. The fine-tuned models are exported and evaluated on the Intel\textregistered{} Movidius\texttrademark{} Myriad\texttrademark{} X VPU, and the resulting on-device metrics are used directly in the NAS fitness loop.
\end{itemize}

Both configurations share the same search space, evolutionary strategy, dataset, and evaluation protocol, enabling a direct comparison of the impact of low-precision-aware training. This setup closely follows prior hardware-aware NAS experiments, with the key distinction that numerical precision effects are now incorporated during optimization rather than deferred to deployment.

\begin{figure*}[t]
    \centering
    \includegraphics[width=\textwidth]{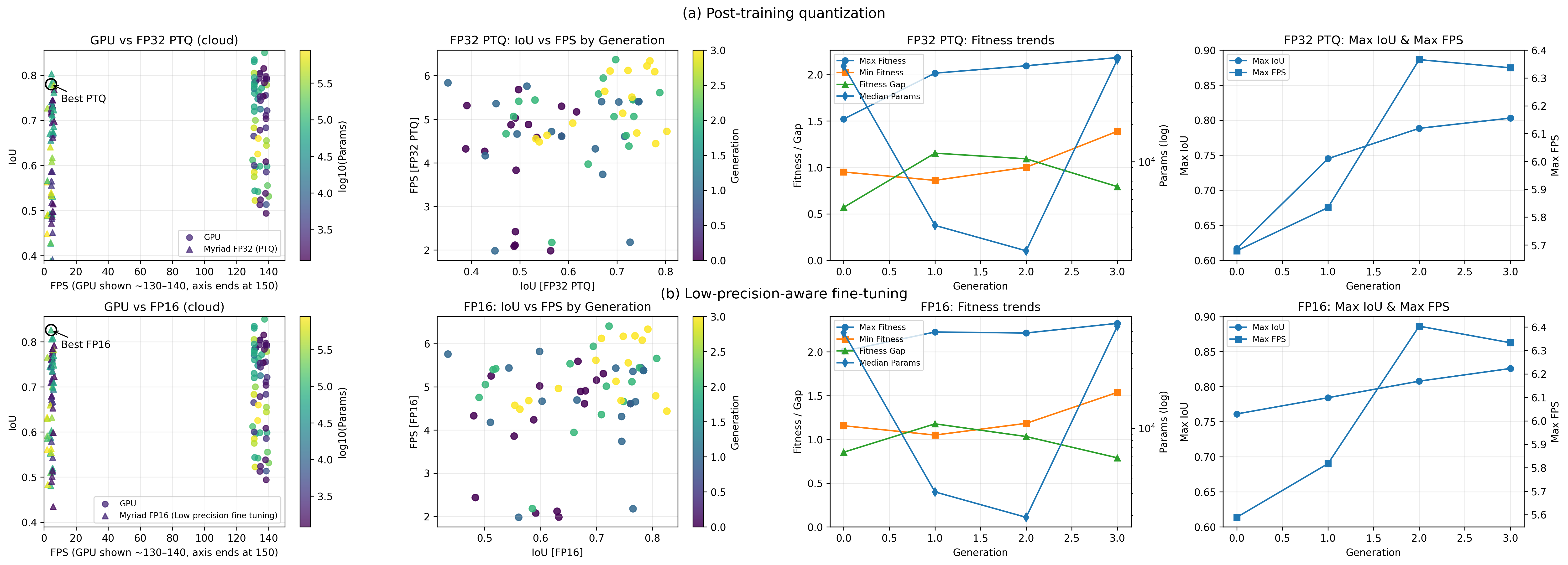}
    \caption{
    Quantitative comparison between post-training precision conversion and low-precision-aware fine-tuning within the hardware-aware NAS framework.
    Top row: PTQ-based optimization. Bottom row: low-precision-aware fine-tuning.
    From left to right: IoU--FPS trade-offs comparing GPU and Myriad VPU execution, IoU versus FPS evolution across generations, fitness trends over generations, and maximum IoU and FPS progression.
    Low-precision-aware training consistently reduces deployment-induced accuracy loss and improves convergence stability on the target device.
    }
    \label{fig:quantitative}
\end{figure*}

\begin{figure*}[t]
    \centering
    \includegraphics[width=\textwidth]{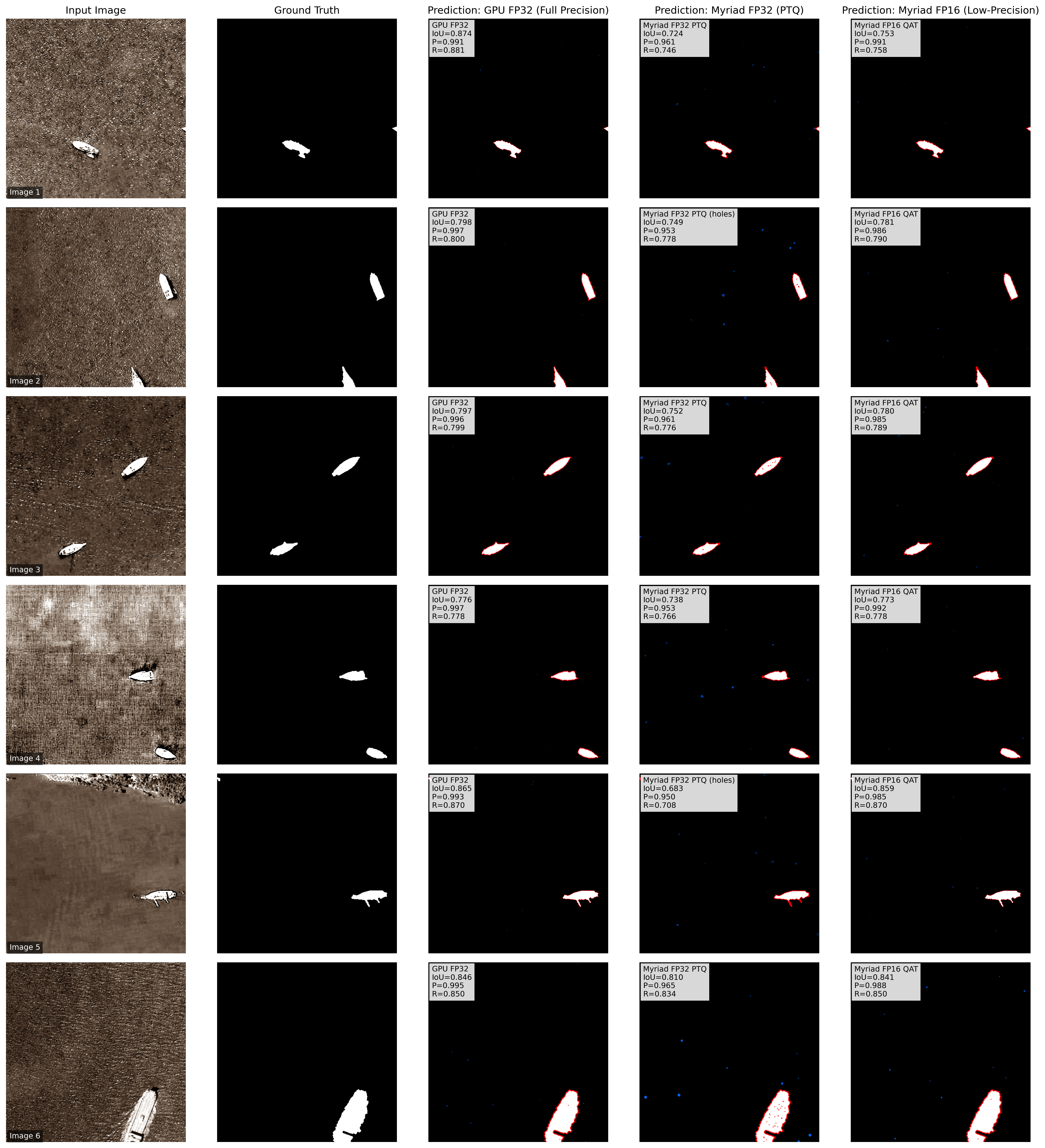}
    \caption{
    Qualitative comparison of vessel segmentation under different numerical conditions.
    From left to right: input image, ground truth mask, GPU FP32 prediction, Myriad X VPU prediction after post-training precision conversion, and Myriad~X VPU prediction after deployment-aligned FP16 training.
    False negatives (missed vessel pixels) are highlighted in \textcolor{red}{red}, while false positives (spurious detections) are highlighted in \textcolor{blue}{blue}.
    Post-training precision conversion introduces fragmented contours, internal holes, and missed detections, particularly for elongated or low-contrast vessels.
    In contrast, deployment-aligned low-precision training preserves vessel connectivity and morphology more reliably under on-device execution.
    }
    \label{fig:qualitative}
\end{figure*}

\subsection{Results}

Figure~\ref{fig:quantitative} summarizes the quantitative behavior of the hardware-aware NAS process under post-training precision conversion and low-precision-aware fine-tuning. The figure reports IoU--FPS trade-offs, fitness evolution across generations, and the progression of maximum accuracy and throughput for both configurations.


Under PTQ, a pronounced discrepancy is observed between GPU-based evaluation and on-device inference. As shown in Fig.~\ref{fig:quantitative}(a), many architectures that achieve high IoU under full-precision GPU evaluation experience a substantial degradation when deployed on Myriad\texttrademark{} X. This effect is particularly evident for architectures with higher parameter counts, indicating sensitivity to deployment-time numerical constraints. While inference throughput remains competitive, the resulting accuracy drop limits the effectiveness of these models for onboard segmentation.

The evolution of fitness values across generations further highlights this limitation. As shown in Fig.~\ref{fig:quantitative}(a), fitness improvements plateau early, and the gap between maximum and median fitness remains relatively large, indicating unstable convergence and a limited number of architectures that jointly satisfy accuracy and efficiency constraints.

In contrast, low-precision-awareness leads to markedly different optimization behavior. As shown in Fig.~\ref{fig:quantitative}(b), architectures optimized under low-precision awareness achieve consistently higher on-device IoU for comparable inference throughput. The IoU--FPS distribution shifts upward, indicating improved robustness to deployment-time numerical effects. Notably, the spread of solutions narrows across generations, suggesting that a larger fraction of the population remains viable under reduced precision.

Fitness trends further confirm this behavior. The maximum fitness increases steadily across generations, while the fitness gap between elite and median architectures decreases, indicating improved population-wide robustness. Importantly, the best-performing architecture achieves a GPU mIoU of approximately 0.84 while maintaining an on-device mIoU close to 0.82 on the Intel\textregistered{} Movidius\texttrademark{} Myriad\texttrademark{} X VPU, significantly reducing the accuracy gap observed under post-training precision conversion.

In addition to quantitative improvements, qualitative inspection reveals clear differences in deployment robustness across numerical configurations.
Figure~\ref{fig:qualitative} compares segmentation outputs produced under GPU FP32 execution, post-training precision conversion on the Intel\textregistered{} Movidius\texttrademark{} Myriad\texttrademark{} X VPU, and deployment-aligned FP16 training.
While GPU FP32 predictions closely match the ground truth masks, post-training precision conversion frequently introduces structural artifacts during on-device execution, including fragmented vessel contours, small holes within predicted regions, and increased false negatives for elongated or low-contrast targets.
These errors are consistent with the numerical sensitivity observed in the quantitative results and reflect the mismatch between optimization-time and deployment-time behavior.

In contrast, models optimized with deployment-aligned low-precision training preserve vessel morphology more reliably under on-device inference.
FP16-aware fine-tuning reduces contour fragmentation, maintains connectivity along vessel hulls, and improves recall for small or partially occluded targets.
Notably, these qualitative gains are achieved without increasing model complexity, confirming that numerical robustness emerges from exposure to deployment-consistent precision during optimization rather than from architectural overparameterization.

Overall, both quantitative metrics and qualitative analysis demonstrate that incorporating low-precision effects directly into the NAS optimization loop substantially improves the alignment between optimization-time behavior and deployment-time performance.
By reducing deployment-induced accuracy loss and preserving segmentation structure under on-device execution without modifying the search space or evolutionary strategy the proposed approach yields architectures that are both efficient and reliable for real-time onboard vessel segmentation.


\section{Conclusion}

This work extends hardware-aware NAS by explicitly aligning optimization-time training with numerical conditions encountered at deployment on low-precision edge accelerators. By integrating deployment-aligned FP16-aware training directly into the NAS evaluation loop, we reduce this mismatch without modifying the search space or evolutionary strategy.

Experiments on vessel segmentation demonstrate that exposure to deployment-consistent numerical effects during optimization substantially improves on-device robustness. On Intel\textregistered{} Movidius\texttrademark{} Myriad\texttrademark{} X, low-precision–aware optimization recovers a large fraction of the accuracy lost under post-training precision conversion, while preserving compact model size and real-time inference performance.

Beyond the specific hardware platform studied, these results highlight a broader methodological insight: architectural efficiency alone is insufficient for reliable edge deployment if numerical behavior is ignored during optimization. For spaceborne and other safety-critical edge AI systems, training pipelines must reflect deployment-time precision constraints to avoid brittle performance in operation.

Future work will explore extending this approach to integer-only and mixed-precision accelerators, as well as jointly optimizing numerical robustness, energy efficiency, and latency within a unified hardware-aware NAS framework. Ultimately, by unifying architecture search, hardware feedback, and numerical-awareness, this work contributes toward a more deployment-centric view of neural optimization, one that is essential for reliable, autonomous intelligence at the edge of space.
{
    \small
    \bibliographystyle{unsrtnat}
    \bibliography{main}
}

\end{document}